\documentclass[lettersize,journal]{IEEEtran}
\usepackage{amsmath,amsfonts,amssymb}
\usepackage{array}
\usepackage[caption=false,font=normalsize,labelfont=sf,textfont=sf]{subfig}
\usepackage{textcomp}
\usepackage{stfloats}
\usepackage{url}
\usepackage{graphicx}
\usepackage{cite}
\usepackage{booktabs}
\usepackage{multirow}
\usepackage{makecell}
\usepackage[hidelinks]{hyperref}
\begin{document}

\title{Spectral-Aware Proposal Enhancement for Diffusion-Based Multispectral UAV Multi-Object Tracking}

\author{Xu~Lin,
        Tingfa~Xu,
        Tianhao~Li,
        Haolin~Qin,
        Huawei~Gao,
        Jichuan~Xing,
        and~Jianan~Li%
\thanks{The authors are with Beijing Institute of Technology, Beijing 100081, China. Tingfa Xu is also with Beijing Institute of Technology Chongqing Innovation Center, Chongqing 401135, China.}
\thanks{\copyright~2026 IEEE. Personal use of this material is permitted. Permission from IEEE must be obtained for all other uses, in any current or future media, including reprinting/republishing this material for advertising or promotional purposes, creating new collective works, for resale or redistribution to servers or lists, or reuse of any copyrighted component of this work in other works. Accepted for publication in \emph{IEEE Geoscience and Remote Sensing Letters}, DOI: 10.1109/LGRS.2026.3742763. This is the author's accepted version; the final published version is available on IEEE Xplore.}
}

\markboth{IEEE Geoscience and Remote Sensing Letters}%
{Author \MakeLowercase{\textit{et al.}}: Spectral-Aware Proposal Enhancement for Diffusion-Based Multispectral UAV Multi-Object Tracking}

\maketitle

\begin{abstract}
Multispectral UAV multi-object tracking (MOT) with oriented bounding boxes remains challenging due to small targets, cluttered backgrounds, and unstable inter-frame associations. Although diffusion-based MOT provides an effective iterative proposal refinement paradigm, existing diffusion-based trackers lack dedicated designs for exploiting multispectral cues in proposal initialization and confidence calibration, limiting tracking robustness in complex UAV scenarios. This letter proposes DSATrack, a diffusion-guided spectral-aware tracking framework for multispectral UAV MOT. Specifically, a lightweight spectral encoder injects multispectral information into the RPN branch through feature fusion, enabling the RPN head to generate high-quality object-aware proposals. These proposals are then injected into the initial proposal set to provide reliable spatial priors for diffusion refinement. Furthermore, a spectral-saliency confidence modulation strategy recalibrates refined tracking confidence according to local multispectral saliency contrast before score thresholding, enhancing spectrally reliable candidates while suppressing unreliable ones. Without modifying the original diffusion refinement head, DSATrack improves both proposal quality and post-refinement candidate selection. Experiments on the MMOT benchmark demonstrate that DSATrack outperforms the adapted DiffusionTrack baseline by 4.6 HOTA, 9.0 MOTA, and 7.7 IDF1 points, showing its effectiveness for multispectral UAV MOT.
\end{abstract}


\section{Introduction}

\IEEEPARstart{M}{ultispectral} unmanned aerial vehicle (UAV) imagery provides complementary spectral information beyond conventional RGB images~\cite{lsa_spectral_dl,lsa_bess,lsa_hsi_rs}, improving target--background separability under degraded appearance conditions, and compact computational spectral imagers make such data increasingly practical to acquire on airborne platforms~\cite{elight_spectropol,lam_spectrometer,elight_meta_imaging}. This is particularly valuable for UAV-view multi-object tracking (MOT), where small, dense, arbitrarily oriented targets, camera motion, weak appearance cues, and cluttered backgrounds cause inaccurate localization, missed detections, and unstable identity association~\cite{lss_tracking}. Reliable multispectral UAV MOT therefore requires both accurate oriented-box localization and robust inter-frame association.

Existing MOT methods fall into tracking-by-detection~\cite{sort,bytetrack,ocsort,botsort}, query-based tracking~\cite{motr,motrv2,memotr,motip}, and diffusion-based proposal refinement. The first detects objects per frame and then associates them, so it depends heavily on detector quality and association rules and is sensitive to missed or unreliable detections in small-target UAV scenes; in the second, the query-object correspondence becomes unstable when targets are small, dense, or weakly distinguishable. Diffusion-based detection instead formulates localization as an iterative denoising process that refines noisy proposals into target boxes~\cite{diffusiondet}, and DiffusionTrack extends this paradigm to MOT by refining paired proposals for joint localization and association~\cite{diffusiontrack}. Such refinement suits UAV-view tracking because it maintains multiple hypotheses and gradually corrects imperfect initialization.
Related efforts improve detection or tracking under specific sensing conditions by strengthening per-frame features or association cues, such as visible--infrared fusion~\cite{vi_tracking}, disparity-guided multiscale fusion~\cite{dmf}, inter-domain adaptive detection~\cite{wasserstein_da}, roadside vehicle tracking~\cite{isn_roadside}, and SAR or sonar target tracking~\cite{tfst,yoloshiptracker,sonar_mot}. None of them acts on the two stages that are specific to diffusion-based trackers: the initial proposal set is still built from tracklets and random noise without image-conditioned priors, and refined candidates are selected purely by learned scores without verification.

To address this gap, this letter proposes DSATrack, a diffusion-guided spectral-aware framework for multispectral UAV MOT. Unlike DiffusionTrack, which initializes proposals from tracklets and noise and trusts refined scores directly, DSATrack introduces spectral cues before and after iterative refinement without redesigning the refinement head: a lightweight spectral encoder with residual-style gated fusion is incorporated into the RPN branch to generate multispectral-enhanced proposals that are injected into the diffusion proposal set as spatial priors, and a spectral-saliency confidence modulation recalibrates refined tracking confidence before score thresholding.

The main contributions of this letter are summarized as follows:

\begin{itemize}

\item We propose DSATrack, which improves proposal initialization and post-refinement candidate selection for multispectral UAV MOT without modifying the diffusion refinement head.

\item We introduce a spectral-aware proposal generation and injection strategy that feeds multispectral information into the RPN branch, providing more reliable object-aware proposals for diffusion refinement.

\item We design a spectral-saliency confidence modulation strategy that recalibrates tracking confidence before score thresholding, improving candidate selection in cluttered UAV scenes.

\end{itemize}

\begin{figure*}[t]
    \centering
    \includegraphics[width=\textwidth]{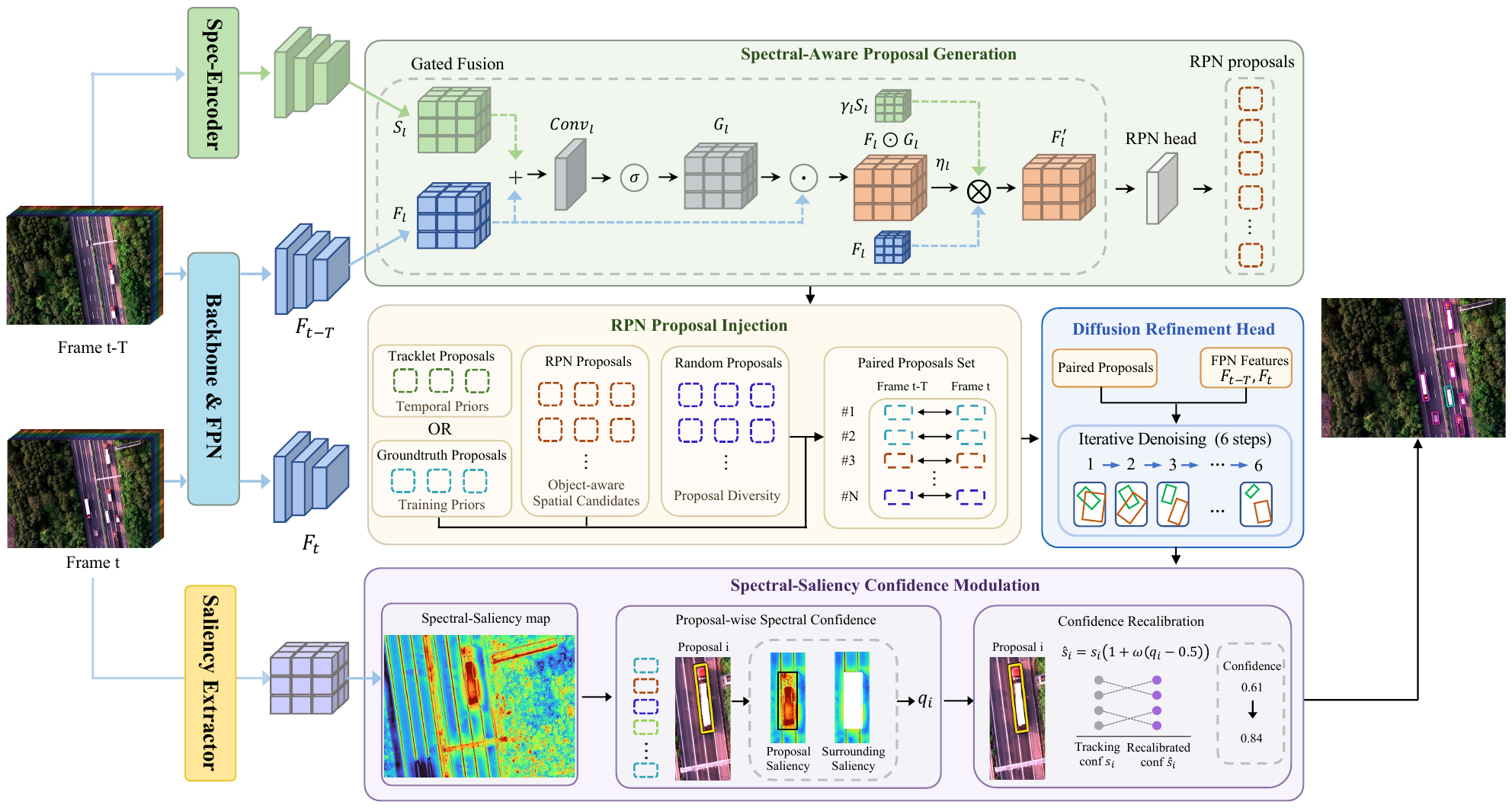}
    \caption{Overall framework of DSATrack. Spectral-aware RPN proposals are injected into the diffusion proposal set, and refined scores are recalibrated by spectral saliency before the final output.}
    \label{fig:framework}
\end{figure*}

\section{Methodology}
\label{sec:methodology}

\subsection{Overview}
\label{subsec:overview}

The overall framework of DSATrack is illustrated in Fig.~\ref{fig:framework}. Given a pair of frames from a multispectral UAV video, the backbone extracts multi-scale pyramid features from both frames; during training the pair is the current frame and a reference frame randomly sampled within a temporal window of $\pm T$ frames, and during inference it is the previous and the current frame. Meanwhile, a lightweight spectral encoder extracts spectral-spatial representations from the multispectral input, which are fused with the pyramid features and fed into an RPN branch~\cite{fasterrcnn} to generate object-aware proposals.
Tracklet, RPN, and random proposals are then assembled into the initial proposal set (Sec.~\ref{subsec:rpn_injection}), the diffusion head progressively updates it to predict oriented bounding boxes with classification and association scores, and the refined scores are recalibrated using multispectral saliency before score thresholding.

\subsection{Spectral-Aware Proposal Generation}
\label{subsec:spectral_rpn}

Injected RPN proposals provide the image-conditioned spatial priors of the initial proposal set, but a standard RPN relies on backbone features that may not preserve discriminative spectral-spatial responses in multispectral UAV imagery~\cite{essfnet}. DSATrack therefore fuses compact spectral-spatial representations with the pyramid features before the RPN head.

Given an 8-channel multispectral image $\mathbf{I}\in\mathbb{R}^{8\times H\times W}$, the spectral encoder performs band mixing with a $1\times1$ convolution, extracts lightweight spatial responses with a depthwise separable block (a depthwise $3\times3$ convolution followed by a pointwise $1\times1$ convolution), recalibrates spectral-channel responses with a channel gate to emphasize informative bands, and builds a multi-scale spectral pyramid with progressive stride-2 convolutions. Let $\mathbf{F}_{l}$ and $\mathbf{S}_{l}$ denote the FPN feature and the corresponding spectral feature at the $l$-th pyramid level.

Before fusion, $\mathbf{S}_{l}$ is spatially aligned with $\mathbf{F}_{l}$ and projected to the same channel dimension through a $1\times1$ convolution. A gated fusion module is then adopted at each pyramid level to adaptively inject spectral information into the proposal branch:
\begin{equation}
\mathbf{G}_{l}
=
\sigma
\left(
\mathrm{Conv}_{l}
\left(
[\mathbf{F}_{l}, \mathbf{S}_{l}]
\right)
\right),
\label{eq:fusion_gate}
\end{equation}
where $[\cdot,\cdot]$ denotes channel-wise concatenation, $\mathrm{Conv}_{l}(\cdot)$ denotes the level-specific convolutional block at the $l$-th pyramid level, and $\sigma(\cdot)$ denotes the sigmoid function. The fused feature is computed as
\begin{equation}
\mathbf{F}'_{l}
=
\mathbf{F}_{l}
+
\lambda_{l}
\left(
\mathbf{G}_{l}
\odot
\mathbf{F}_{l}
\right)
+
\mu_{l}
\mathbf{S}_{l},
\label{eq:spectral_fusion}
\end{equation}
where $\odot$ denotes element-wise multiplication and the coefficients $\lambda_l$ and $\mu_l$ are learnable weights made non-negative by the Softplus function, so that the fusion strength is learned at each pyramid level.

The fused pyramid features are fed into a standard RPN head~\cite{fasterrcnn} that predicts objectness scores and box offsets. Since the branch only provides coarse spatial priors, it predicts horizontal proposals for efficiency; after decoding and post-processing, the retained proposals are converted into five-dimensional oriented proposals by assigning the same constant initial angle to all of them, and their orientation is re-estimated by the refinement head.

\subsection{RPN-Guided Proposal Injection}
\label{subsec:rpn_injection}

In the original diffusion tracking pipeline, active trajectories from previous frames are propagated as tracklet proposals, which serve as temporal priors for continuing targets but are unavailable for newly appearing or missed ones. DSATrack therefore complements them with spectral-aware RPN proposals that provide image-conditioned spatial priors, while random proposals preserve diversity; the three sources form a fixed-size initial proposal set.

For pairwise tracking, proposals with the same index in two frames are treated by the refinement head as the same target hypothesis. To avoid index mismatch caused by independently generated RPN proposals, we generate RPN proposals from the previous frame and reuse them for the current-frame proposal set, which provides index-aligned initial proposals rather than assuming static targets; inter-frame motion, scale variation, and rotation are handled by iterative refinement.
During training, ground-truth boxes take the place of tracklet proposals as supervised references, while RPN and random proposals are retained; this is the only difference between the training and inference proposal sets.

\subsection{Spectral-Saliency Confidence Modulation}
\label{subsec:spectral_confidence_modulation}

After refinement, the model outputs a fixed number of oriented boxes with classification and association confidence scores. Since these scores are predicted from deep features, background-like candidates may still obtain high confidence when local multispectral evidence is weak, so DSATrack recalibrates them using local spectral saliency before score thresholding.

To avoid an additional trainable verification branch, a single-channel spectral saliency map is computed by aggregating multi-scale Gaussian difference responses over all spectral channels:
\begin{equation}
\mathbf{S}(x,y)
=
\operatorname{Norm}
\left(
\sum_{\sigma \in \Omega}
\sum_{c=1}^{C}
\left|
\mathbf{I}_{c}(x,y)
-
\mathcal{G}_{\sigma}
(\mathbf{I}_{c})(x,y)
\right|
\right),
\label{eq:saliency_map}
\end{equation}
where $\mathcal{G}_{\sigma}(\cdot)$ denotes Gaussian filtering and $\operatorname{Norm}(\cdot)$ min-max normalization over the whole frame, with the scale set $\Omega=\{2,4\}$. %

Given the classification and association confidences $p_{\mathrm{cls}}^{i}$ and $p_{\mathrm{assoc}}^{i}$ of the $i$-th refined proposal, the original tracking confidence is
\begin{equation}
s_i =
\sqrt{
p_{\mathrm{cls}}^{i}
p_{\mathrm{assoc}}^{i}
}.
\label{eq:tracking_score}
\end{equation}

For each refined box, the mean foreground saliency $\bar{S}_{\mathrm{fg}}^{\,i}$ is computed inside the box and the background statistics $\mu_{\mathrm{bg}}^{\,i}$ and $\sigma_{\mathrm{bg}}^{\,i}$ from the ring between the box and its $1.5\times$ enlarged box with the same orientation, giving the spectral confidence
\begin{equation}
q_i =
\operatorname{sigmoid}
\left(
\frac{
\bar{S}_{\mathrm{fg}}^{\,i}
-
\mu_{\mathrm{bg}}^{\,i}
}{
\sigma_{\mathrm{bg}}^{\,i}
+
\epsilon
}
\right).
\label{eq:spectral_conf}
\end{equation}

The constant $\epsilon=10^{-6}$ avoids division by zero. Finally, the tracking confidence is recalibrated as
\begin{equation}
\hat{s}_i =
s_i
\left(
1 + w(q_i - 0.5)
\right),
\label{eq:score_recalibration}
\end{equation}
where $w$ controls the modulation strength. It is applied to the candidates that pass the original score threshold, keeping at most the 500 highest-scoring ones, and $\hat{s}_i$ is thresholded again before class-wise rotated non-maximum suppression (NMS), so that proposals with stronger local spectral support are preferentially retained.

\section{Experiments}
\label{sec:experiments}

\subsection{Experimental Settings}
\label{subsec:exp_settings}

Experiments follow the official split and protocol of the MMOT benchmark~\cite{mmot}, which provides 75 training and 50 testing 8-channel UAV-view sequences with oriented-box annotations, about 13.8K frames and 488.3K object annotations in total. We report HOTA, MOTA, IDF1, DetA, and AssA under both class-averaged and detection-averaged settings.
YOLO11l~\cite{yolo11} is used as the backbone, initialized with the public weights released for MMOT. DiffusionTrack~\cite{diffusiontrack} is adapted to this setting by using the same 8-channel backbone and predicting oriented boxes with rotated-IoU matching and rotated NMS, all other components being unchanged; this adapted model is the baseline.
The diffusion head consists of six cascaded refinement stages with a single sampling step and uses 500 proposals in training and 1000 in inference, including at most 64 RPN proposals retained with a score threshold of 0.3 and an NMS threshold of 0.7. The detection and tracking stages are each trained for 30 epochs with a batch size of 4 on four RTX 3090 GPUs, using AdamW with a weight decay of $10^{-4}$ and a peak learning rate of $6.25\times10^{-5}$ with one warm-up epoch and cosine decay; Mosaic and MixUp are disabled in the last five epochs, and the temporal window is $T=3$. At inference, the detection and association thresholds are 0.50 and 0.20, the rotated NMS threshold is 0.60, the re-association interval is 5 frames, and $w=0.5$.

\subsection{Experimental Results and Analysis}
\label{subsec:results_analysis}

Following the MMOT benchmark setting, we compare DSATrack with representative tracking-by-detection, tracking-by-query, and tracking-by-diffusion MOT methods under the MSI input setting.

\begin{table*}[t]
\centering
\caption{Comparison of representative MOT algorithms with MSI input on MMOT. $^{\dagger}$Adapted to 8-channel input and oriented boxes with the YOLO11l backbone (Sec.~\ref{subsec:exp_settings}).}
\label{tab:sota_comparison}
\renewcommand{\arraystretch}{1.05}
\setlength{\tabcolsep}{4.2pt}
\setlength{\aboverulesep}{0.25ex}
\setlength{\belowrulesep}{0.25ex}
\resizebox{\textwidth}{!}{%
\begin{tabular}{
>{\centering\arraybackslash}m{1.45cm}
!{\vrule width 0.4pt}
l
!{\vrule width 0.4pt}
ccccc
!{\vrule width 0.4pt}
ccccc
}
\toprule
\multirow{2}{*}{Type}
& \multirow{2}{*}{Method}
& \multicolumn{5}{c}{Class-Averaged}
& \multicolumn{5}{c}{Detection-Averaged} \\
&
& HOTA & MOTA & IDF1 & DetA & AssA
& HOTA & MOTA & IDF1 & DetA & AssA \\
\midrule

\multirow{4}{*}{\makecell[c]{Tracking by\\Detection}}
& SORT~\cite{sort}      & 27.2 & 24.3 & 29.1 & 25.7 & 30.0 & 35.0 & 25.7 & 33.7 & 27.6 & 44.8 \\
& ByteTrack~\cite{bytetrack} & 40.5 & 34.2 & 44.1 & 37.0 & 46.2 & 46.0 & 37.8 & 46.7 & 41.9 & 51.5 \\
& OC-SORT~\cite{ocsort}   & 29.5 & 25.1 & 31.9 & 27.3 & 32.8 & 37.5 & 27.5 & 37.0 & 29.5 & 48.0 \\
& BoT-SORT~\cite{botsort}  & \textbf{53.6} & \textbf{46.2} & \textbf{61.0} & \textbf{45.7} & 64.6
            & \textbf{60.7} & \textbf{59.4} & \textbf{69.4} & \textbf{55.0} & 68.7 \\
\midrule

\multirow{4}{*}{\makecell[c]{Tracking by\\Query}}
& MOTR~\cite{motr}   & 39.0 & 26.5 & 44.6 & 27.1 & 60.1 & 48.4 & 32.2 & 54.7 & 35.4 & 68.4 \\
& MOTRv2~\cite{motrv2} & 49.2 & 43.1 & 57.3 & 37.8 & \textbf{67.7} & 54.5 & 50.9 & 64.6 & 44.1 & 68.8 \\
& MeMOTR~\cite{memotr} & 42.3 & 31.3 & 45.9 & 29.3 & 66.3 & 50.9 & 40.8 & 56.0 & 37.1 & \textbf{70.9} \\
& MOTIP~\cite{motip}  & 39.0 & 28.8 & 43.9 & 33.8 & 49.6 & 43.1 & 37.3 & 46.3 & 43.7 & 43.8 \\
\midrule

\multirow{2}{*}{\makecell[c]{Tracking by\\Diffusion}}
& DiffusionTrack$^{\dagger}$~\cite{diffusiontrack}       & 47.9 & 31.4 & 51.3 & 41.4 & 58.1 & 50.7 & 35.5 & 51.8 & 47.2 & 56.9 \\
& DSATrack (Ours)      & 52.5 & 40.4 & 59.0 & 43.8 & 65.5 & 55.9 & 44.9 & 61.0 & 48.6 & 66.5 \\
\bottomrule
\end{tabular}%
}
\end{table*}

\begin{figure*}[t]
    \centering
    \includegraphics[width=\textwidth]{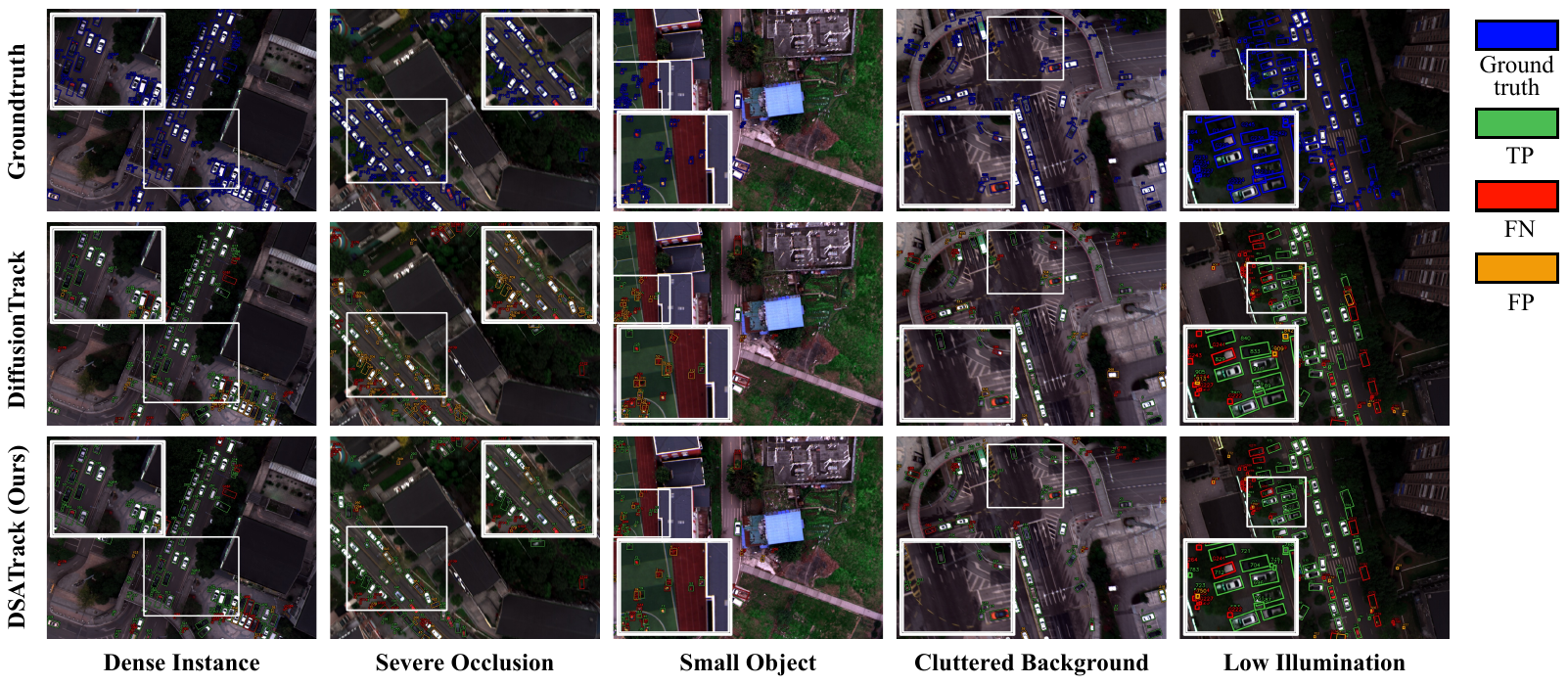}
    \caption{Qualitative comparison of spatial tracking results.}
    \label{fig:spatial_visualization}
\end{figure*}

\begin{figure*}[t]
    \centering
    \includegraphics[width=0.92\textwidth]{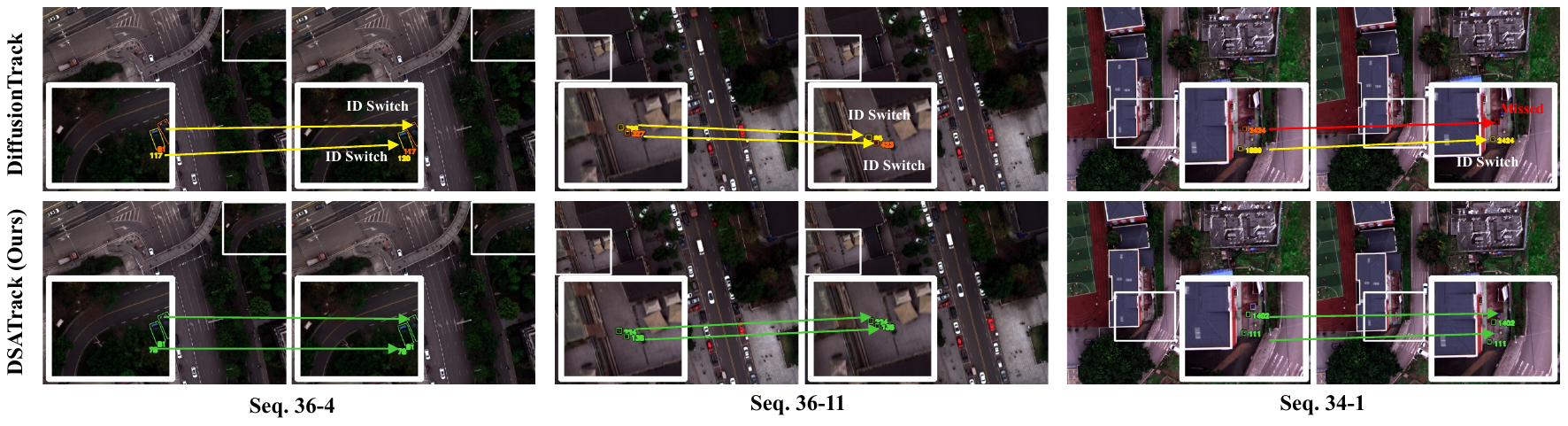}
    \caption{Qualitative comparison of temporal association results on video clips.}
    \label{fig:temporal_visualization}
\end{figure*}

As shown in Table~\ref{tab:sota_comparison}, DSATrack does not outperform all competing methods: BoT-SORT, which relies on a strong separately trained detector, achieves higher HOTA, MOTA, and IDF1. Within the tracking-by-diffusion paradigm, however, it clearly improves the adapted baseline, raising HOTA, MOTA, and IDF1 by 4.6, 9.0, and 7.7 points under the class-averaged setting. The qualitative results in Fig.~\ref{fig:spatial_visualization} and Fig.~\ref{fig:temporal_visualization} further show more complete target coverage, more accurate oriented boxes, and more stable identities in cluttered scenes.

With an $800\times1440$ input on one RTX 3090 GPU, DSATrack has 272.5\,M parameters (+2.6\% over the baseline, all in the spectral-aware RPN branch), adds 142.1 GFLOPs to the 230.2 GFLOPs of the backbone per frame pair, and runs at 1.65 FPS versus 1.70 FPS; the modulation is parameter-free and takes 1.2\,ms per frame.

\begin{figure}[t]
    \centering
    \includegraphics[width=\columnwidth]{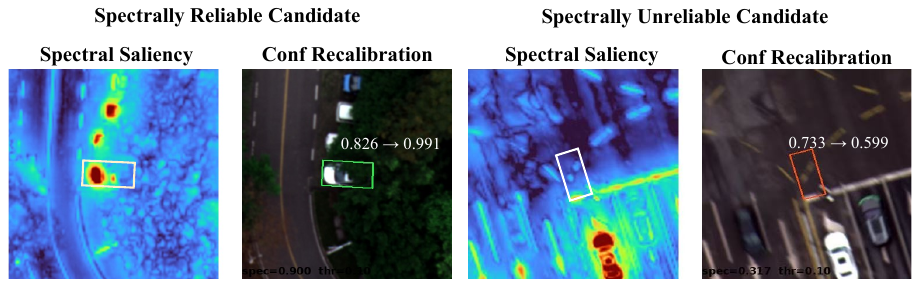}
    \caption{Visualization of spectral-saliency confidence modulation before score thresholding.}
    \label{fig:score_modulation_visualization}
\end{figure}

Fig.~\ref{fig:score_modulation_visualization} illustrates that the modulation increases the scores of candidates with strong local spectral contrast and suppresses unreliable ones before thresholding.
\subsection{Ablation Study}
\label{subsec:ablation}

\begin{table}[t]
\centering
\caption{Ablation of the proposed components. A, B, and C denote RPN-guided proposal injection, spectral-aware RPN, and spectral-saliency confidence modulation, respectively. Without B, the RPN of A takes a pseudo-RGB input (bands 1, 2, and 4).}
\label{tab:ablation}
\renewcommand{\arraystretch}{1.0}
\setlength{\tabcolsep}{4.5pt}
\small
\begin{tabular}{ccc|ccccc}
\hline
A & B & C & HOTA & MOTA & IDF1 & DetA & AssA \\
\hline
$\times$ & $\times$ & $\times$ & 47.9 & 31.4 & 51.3 & 41.4 & 58.1 \\
$\checkmark$ & $\times$ & $\times$ & 49.6 & 35.1 & 54.6 & 41.7 & 61.4 \\
$\times$ & $\times$ & $\checkmark$ & 49.8 & 35.3 & 54.0 & 42.8 & 60.4 \\
$\checkmark$ & $\checkmark$ & $\times$ & 51.1 & 37.8 & 56.5 & 43.5 & 62.2 \\
$\checkmark$ & $\checkmark$ & $\checkmark$ & 52.5 & 40.4 & 59.0 & 43.8 & 65.5 \\
\hline
\end{tabular}
\normalsize
\end{table}

As shown in Table~\ref{tab:ablation}, each component improves over the baseline: RPN-guided injection provides object-aware initialization, the spectral-aware RPN further improves DetA with more reliable spatial priors, and the confidence modulation, which is also effective on its own, brings additional gains in IDF1 and AssA by suppressing unstable candidates before thresholding.

\begin{table}[t]
\centering
\caption{Ablation on the number of injected RPN proposals.}
\label{tab:rpn_number}
\renewcommand{\arraystretch}{1.0}
\setlength{\tabcolsep}{4.5pt}
\small
\begin{tabular}{c|ccccc}
\hline
\#RPN & HOTA & MOTA & IDF1 & DetA & AssA \\
\hline
32  & 52.29 & 40.42 & 58.58 & 43.78 & 64.85 \\
64  & 52.53 & 40.44 & 58.98 & 43.76 & 65.53 \\
128 & 52.50 & 40.36 & 58.91 & 43.76 & 65.43 \\
256 & 52.12 & 40.36 & 58.10 & 43.87 & 64.40 \\
\hline
\end{tabular}
\normalsize
\end{table}
The number of injected RPN proposals controls the balance between object-aware initialization and proposal diversity. As shown in Table~\ref{tab:rpn_number}, HOTA, IDF1, and AssA peak at 64 proposals and degrade at 256, although DetA slightly increases, suggesting that excessive proposals introduce redundant or less reliable candidates; 64 is therefore adopted in all experiments.

\section{Conclusion}
\label{sec:conclusion}

This letter presented DSATrack, which improves diffusion-based multispectral UAV MOT without modifying the refinement head. Improving only how proposals are initialized and how refined candidates are selected raises HOTA by 4.6 points, and since the confidence modulation needs no additional parameters and costs 1.2\,ms per frame, it can be attached to existing diffusion-based trackers on resource-constrained UAV platforms. DSATrack is still less effective on very small targets, whose local saliency contrast is unreliable, and on targets spectrally similar to their background, and remains below tracking-by-detection methods with strong detectors; handling these cases and reducing the cost of the inherited refinement head are left for future work.

\end{document}